\pdfoutput=1

\documentclass[11pt]{article}

\usepackage{EMNLP2023}

\usepackage{times}
\usepackage{latexsym}

\usepackage[T1]{fontenc}

\usepackage[utf8]{inputenc}

\usepackage{microtype}

\usepackage{inconsolata}

\usepackage{graphicx}
\usepackage{tabularx}
\usepackage{makecell}
\usepackage{xcolor}

\def\toolname{SimplyRetrieve}
\def\githubname{RCGAI}

\title{\toolname: A Private and Lightweight Retrieval-Centric Generative AI Tool}

\author{Youyang Ng,
  Daisuke Miyashita,
  Yasuto Hoshi,
  Yasuhiro Morioka, \\
  {\bf Osamu Torii},
  {\bf Tomoya Kodama},
  {\bf Jun Deguchi} \\
  Kioxia Corporation, Japan \\
  \texttt{youyang.ng@kioxia.com} \\}

\begin{document}
\maketitle
\begin{abstract}
Large Language Model (LLM) based Generative AI systems have seen significant progress in recent years. Integrating a knowledge retrieval architecture allows for seamless integration of private data into publicly available Generative AI systems using pre-trained LLM without requiring additional model fine-tuning. Moreover, Retrieval-Centric Generation (RCG) approach, a promising future research direction that explicitly separates roles of LLMs and retrievers in context interpretation and knowledge memorization, potentially leads to more efficient implementation. \emph{\toolname{}} is an open-source tool with the goal of providing a localized, lightweight, and user-friendly interface to these sophisticated advancements to the machine learning community. \toolname{} features a GUI and API based RCG platform, assisted by a Private Knowledge Base Constructor and a Retrieval Tuning Module. By leveraging these capabilities, users can explore the potential of RCG for improving generative AI performance while maintaining privacy standards. The tool is available at \url{https://github.com/\githubname/\toolname} with an MIT license.

\end{abstract}

\section{Introduction}
\label{sec:introduction}

Generative-based Natural Language Processing (NLP) has witnessed significant progress \citep{GPT3} in recent years. With the introduction of Transformer \citep{Transformer} architecture, the possibility of developing high-accuracy language models that can perform tasks such as text generation, text summarization and language translation has become a reality. These models \citep{GPT3,PALM}, when scaled up to billions of parameters \citep{emergent}, have shown remarkable improvements in text generation tasks such as zero-shot inference, popularized the term \emph{Generative AI}. Instead of model fine-tuning, careful design of prompts has proven effective in adapting these models to specific domains for various tasks \citep{GPT3}. This has given rise to the field of prompt-engineering. Additionally, Chain-of-Thought \citep{CoT,kojima2022large} decomposes a complex task assigned into manageable steps, thereby expanding the capabilities of generative-based language models even further.

\begin{figure}[t]
	\centering
  \includegraphics[width=0.8\columnwidth]{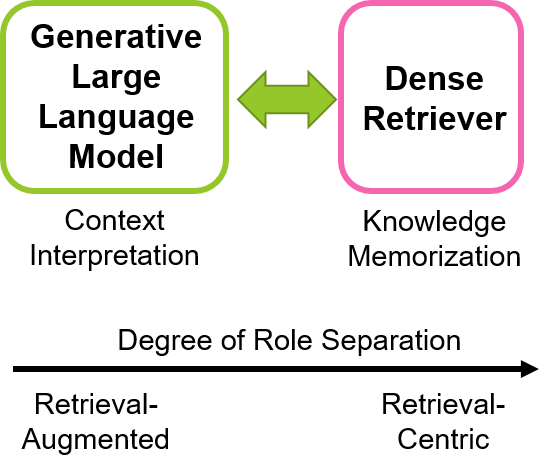}
  \caption{Retrieval-Centric Generation (RCG) approach presents an innovative concept that leverages the mutually beneficial interaction between LLMs and retrievers for more efficient context interpretation and knowledge memorization. Increased clarity in role-separation between context interpretation and knowledge memorization can potentially boost the performance of generative AI systems.}
	\label{Fig.1:retrievalcentric}
\end{figure}

Training large language models (LLMs) requires immense computational resources, often involving thousands of high-end GPUs. Fine-tuning these models can also be challenging. Although prompt-engineering helped to reduce the need for fine-tuning, there was still noticeable instruction misalignment when interacting with a human user. To address this issue, techniques such as reinforcement learning from human feedback (RLHF) \citep{RLHF} have been explored to align the behavior of LLMs with human values \citep{InstructHF,ChatGPT}. Additionally, QLoRA \citep{qlora}, combining low-rank adaptation technique \citep{hu2022lora} and quantization technique, has made it possible to fine-tune these models on individual developer's hardware, making them more accessible to a wider range of users. Despite these advances, there are still limitations to the capacity of LLMs, and they do not inherently recognize information that was not present during training and fine-tuning. Memorization of factual knowledge in the long tail is also a challenge \citep{mallen-etal-2023-trust}.

Most recently, there has been growing interest in integrating external knowledge sources into LLMs for generating text \citep{RETRO, REALM, RAG}. Similar approaches have also been proposed in solving computer vision tasks \citep{knnclip,Iscen_2023_CVPR}. Retrieval-Augmented Generation (RAG) \citep{RAG} architecture is an approach that enhances the capabilities of LLMs by incorporating external data sources using a sparse or dense retriever \citep{karpukhin-etal-2020-dense}, enabling the use of privately owned data without requiring retraining or fine-tuning the LLM \citep{LangChain}. However, developing retrieval-augmented LLM-based generative models is still in its early stages. Our proposed tool can help facilitate these developments.

Additionally, we introduce a new architectural concept called \emph{Retrieval-Centric Generation} (RCG), which builds upon the Retrieval-Augmented Generation approach by emphasizing the crucial role of the LLM in interpreting context and entrusting knowledge memorization to the retriever component, putting greater importance on retriever, as depicted in Figure \ref{Fig.1:retrievalcentric}. By separating context interpretation from knowledge memorization, this approach has the potential to reduce the scale \citep{carlini2023quantifying} of the LLM required for generative tasks, leading to more efficient and interpretable results. Moreover, this approach may help mitigate hallucinations \citep{maynez-etal-2020-faithfulness} by limiting the scope of the LLM's generation. Once we define RCG as above, we can re-define RAG that enables more permissible usage of LLM's inherent knowledge, whereas RCG prioritizes clear demarcations between context interpretation and knowledge memorization.

\toolname{} is an open-source tool aimed at providing a localized, lightweight, and user-friendly interface to Retrieval-Centric Generation approach to the machine learning community. This tool encompasses a GUI and API based RCG platform, assisted by a Private Knowledge Base Constructors and a Retrieval Tuning Module. \toolname{} is designed to be simple and accessible to the community, as well as end-users. Our retrieval-centric platform incorporates multiple selectable knowledge bases featuring Mixtures-of-Knowledge-Bases (MoKB) mode and Explicit Prompt-Weighting (EPW) of retrieved knowledge base. By designing \toolname{} with these features, we enable the machine learning community to explore and develop with a lightweight, private data interface to LLM-based generative AI systems, with a focus on retrieval-centric generation. Potential developments that can be explored using this tool include: (1) examining the effectiveness of retrieval-centric generation in developing safer, more interpretable, and responsible AI systems; (2) optimizing the efficiency of separating context interpretation and knowledge memorization within retrieval-centric generation approach; and (3) improving prompt-engineering techniques for retrieval-centric generation. \toolname{} is available at \url{https://github.com/\githubname/\toolname}.

Our contributions can be summarized as follows:
\begin{itemize}
  \item We propose \toolname{}, an innovative and user-friendly tool that leverages GUI and API platform to facilitate a Retrieval-Centric Generation approach. This platform is further strengthened by two key components: Private Knowledge Base Constructor and Retrieval Tuning Module. 

  \item We open sourced our tool to the machine learning community and identify potential development directions of Retrieval-Centric Generation.
\end{itemize}

\begin{figure*}[t]
	\centering
  \includegraphics[width=\textwidth]{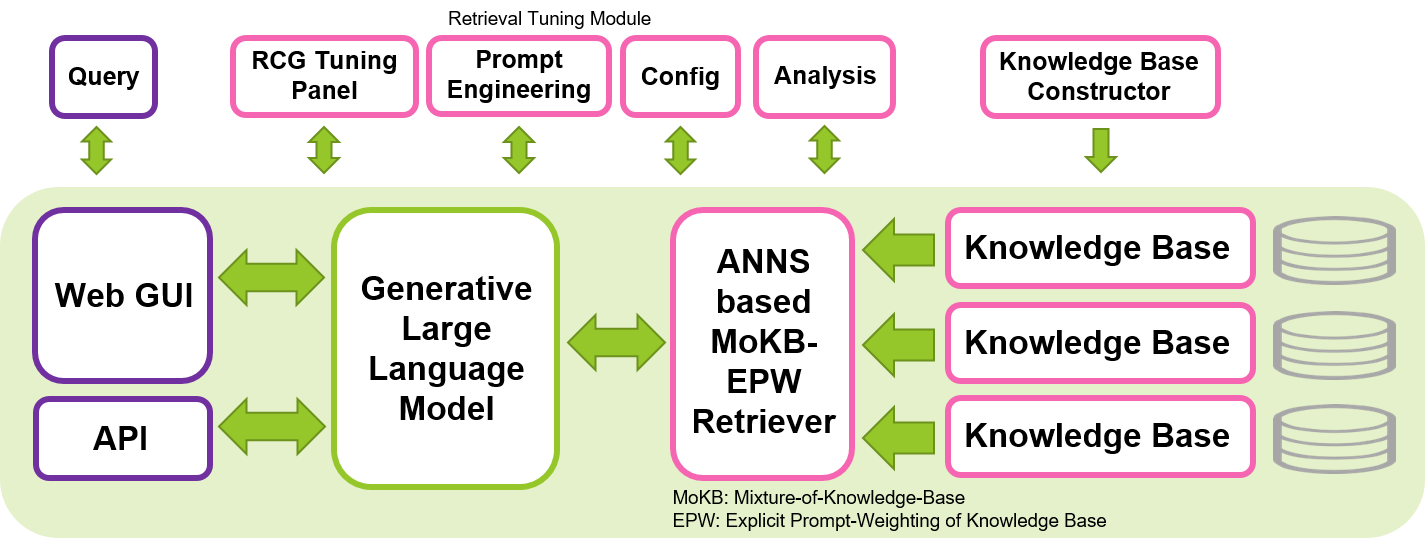}
  \caption{\toolname{} is an open-source tool that provides a localized, lightweight, and user-friendly interface to the Retrieval-Centric Generation approach for the machine learning community. This tool features a GUI and API based RCG platform, assisted by a Private Knowledge Base Constructor and a Retrieval Tuning Module.}
	\label{Fig.2:tooloverview}
\end{figure*}

\section{Related Works}
\label{sec:relatedworks}

The emergence of Retrieval-Augmented Generation architecture has spurred the development of numerous open-source tools. The ChatGPT Retrieval Plugin\footnote{\url{https://github.com/openai/chatgpt-retrieval-plugin}}, for instance, integrates the ability to retrieve and enhance personal or organizational documents into the widely used ChatGPT model \citep{ChatGPT}. Similarly, fastRAG \citep{fastRAG} provides a streamlined platform for constructing efficient retrieval-augmented generation pipelines. Additionally, LangChain \citep{LangChain} offers a comprehensive generative chat AI library featuring agents, data augmentation, and memory capabilities. Finally, Haystack \citep{haystack} presents an all-encompassing NLP framework supporting question answering, answer generation, semantic document search, and retrieval-augmentation. Both LangChain and Haystack employ agent-based pipelining techniques and can process complex queries. However, this complexity may hinder the explainability of LLMs, making it challenging to interpret their performance in retrieval-augmented settings.

On the other hand, our work offers a lightweight and transparent approach to implementing sophisticated retrieval-centric, as well as retrieval-augmented architecture, while maintaining a strong emphasis on response interpretability and wider accessibility to the community. Unlike previous works such as PrivateGPT \citep{privategpt}, which provides a privacy-preserving chat AI tool but lacks customization options and analytical capabilities, our tool offers a comprehensive set of features for tailoring and analyzing retrieval-centric generation.

Furthermore, to the best of our knowledge, we are the first to introduce RCG concept and show initial experiments of it using our tool.

\begin{figure*}[t]
	\centering
  \includegraphics[width=\textwidth]{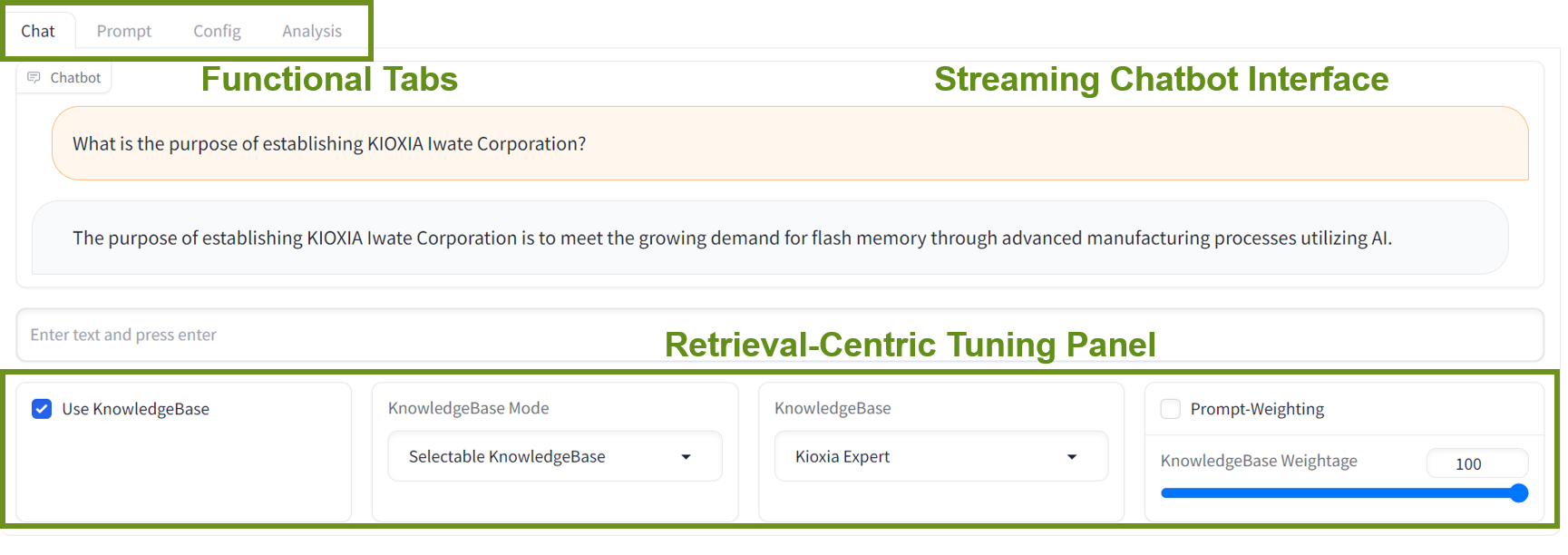}
  \caption{The GUI design of \toolname{} features four primary tabs. The Chat tab serves as the central query and response interface with retrieval-centric tuning panel. The Prompt tab provides an intuitive editor for modifying, updating, and saving prompts used by the AI. The Config tab enables users to customize various tool settings and save their preferences. Finally, the Analysis tab offers a comprehensive analytics platform for analyzing and logging data related to \toolname{}'s performance and usage.}
	\label{Fig.3:guioverview}
\end{figure*}

\section{Tool Design}
\label{sec:tooldesign}

\toolname{} is designed to deploy RCG pipeline: construct knowledge base, tune architecture, make predictions. In this paper, we focus on describing the core specifications of the tool. For details about the setup procedures, refer to the repository of \url{https://github.com/\githubname/\toolname}.

\subsection{GUI and API based Retrieval-Centric Generation Platform}
As shown in Figure \ref{Fig.2:tooloverview}, there are two dense models in our tool: an LLM and an Approximate Nearest Neighbor Search (ANNS) based Knowledge Retriever. The LLM can be any one of the off-the-shelf open-source LLM models available in Hugging Face \citep{huggingface}, ranging from 1B to more than 100B-scale in parameters such as \citet{llama,touvron2023llama}. The Knowledge Retriever employs a dense retriever that is compatible with various embedding models available in Hugging Face. Additionally, our tool allows integration of multiple knowledge bases simultaneously, enabling user-selectable knowledge bases depending on the specific use case.

In terms of the GUI, we have designed a simple yet intuitive layout using Gradio \citep{abid2019gradio}, which provides a familiar streaming chatbot interface with user control for managing the running modes of the retriever, engineering prompts, and configuring the tool. As depicted in Figure \ref{Fig.3:guioverview}, our GUI features a comprehensive \emph{retrieval-centric tuning panel} for functions including \emph{manual knowledge base selection} from multiple sources and \emph{Mixture-of-Knowledge-Base} modes. Moreover, we employ \emph{Explicit Prompt-Weighting} of retrieval to adjust the level of influence exerted by the retriever. To ensure seamless integration, we also developed a comprehensive API access function using the Gradio Client Interface, and we allow multi-user concurrent access to both UIs, leveraging Gradio's queue functionality to manage requests efficiently.

The retrieval-centric tuning panel enables lightweight and simplistic access to RCG. By using the manual knowledge base selection mode, users can construct and import multiple private knowledge bases simultaneously into this tool. The ability to select the most relevant knowledge base for each task allows users to maintain control over the selection process while avoiding any unexpected outcomes. Our MoKB mode enables automatic selection of the most suitable knowledge base based on the similarity between the query and knowledge base functional descriptions. We use semantic cosine similarity of embedding space to calculate these scores, providing an efficient and lightweight approach to knowledge base auto-selection. By updating the functional descriptions in the configuration file, users can further enhance the accuracy of the selection algorithm.

Additionally, our Explicit Prompt-Weighting feature allows manual adjustment of the degree of influence of retrievers on the language model, enabling customized control over the balance between retriever and LLM. Through prompt-engineering or token weight adjustment, users can adapt the tool to their specific needs, ensuring optimal performance. \toolname{} has incorporated Explicit Prompt-Weighting through prompt-engineering, where the weightage can be adjusted to fine-tune the percentage of knowledge tokens to be used in the prompt out of retrieved tokens. However, we have not implemented token weight adjustment in this study and leave it for future work.

\subsection{Private Knowledge Base Constructor}
Our Retrieval-Centric Generation Platform is assisted by a Private Knowledge Base Constructor that creates a local and personalized knowledge base using the user's documents. This constructor employs a scalable documents loader that can handle large volumes of documents by chunking and streaming the loading, splitting and knowledge base creation processes, allowing for efficient document processing. The constructor supports various document formats such as PDF, TXT, DOC, DOCX, PPT, PPTX, HTML, MD, CSV, among others, and can be easily expanded by editing configuration file. Additionally, the length of passages in the documents splitting function is easily configurable to meet specific requirements.

After generating the sources for the knowledge base, we use a dense encoder to convert the text into numerical embeddings that can be used for semantic search and retrieve. To accommodate large-scale knowledge bases, we utilize ANNS for efficient semantic retrieval. By default, our tool employs the Hierarchical Navigable Small Worlds (HNSW) \citep{10.1109/TPAMI.2018.2889473} algorithm, but we also provide support for flat indexing and the IVFPQ-HNSW method, which combines inverted file indexing with product quantization and HNSW course quantizers. The Index Constructor function automatically creates the required index files for semantic searching. We implement our indexing function by using Faiss library \citep{johnson2019billion}.

\subsection{Retrieval Tuning Module}
The Retrieval Tuning Module of our tool includes three key functionalities: prompt-engineering, tool configuration, and analysis and data logging. The prompt-engineering functionality allows users to easily edit, update, and save retrieval-related prompts using a user-friendly Prompt Tab within our GUI. Available prompts are \emph{AI Prefix, Retriever Prefix, Retriever Suffix, Model Prefix and Model Suffix}. The configuration functionality enables users to modify and save all configurable settings via the Config Tab within our GUI. Finally, the analysis and data logging functionality collects and displays retrieval-related analysis data, including retrieved knowledge base, query, response, sentence-level and token-level similarity scores, in the Analysis Tab of our GUI. Similarity scores are calculated based on both semantic cosine similarity of sentence-to-sentence embeddings and all-token-to-token embeddings. This approach allows us to capture both local and global similarities between sentences, leading to more accurate assessments of their comparability. Additionally, users can save all logged data to a log file for further analysis. GUI designs are depicted in Figure \ref{Fig:guiprompt}, \ref{Fig:guiconfig} and \ref{Fig:guianalysis} of Appendix \ref{sec:applications}. To deploy an end-user mode, users can simply disable the update functions in the Retrieval Tuning Module through command-line options.

\section{Evaluations}
\label{sec:evaluations}

In this section, we perform several qualitative evaluations to demonstrate the usability and behavior of our tool. We construct our knowledge base using the most recent information available on the website of an organization\footnote{\url{https://www.kioxia.com/en-jp/top.html}}. We utilize the models publicly available on Hugging Face, \emph{Wizard-Vicuna-13B}\footnote{\url{https://huggingface.co/ehartford/Wizard-Vicuna-13B-Uncensored}} \citep{xu2023wizardlm,vicuna2023} as the LLM and \emph{Multilingual-E5-base}\footnote{\url{https://huggingface.co/intfloat/multilingual-e5-base}} \citep{wang2022text} as the encoder for our evaluations, unless specified otherwise. We load both models into a single Nvidia A100 GPU in 8-bit INT8 mode for lower memory usage and higher inference speed. We set temperature of LLM to 0. We utilize HNSW for indexing of knowledge bases and set the number of passages retrieved to 5.

\subsection{Qualitative Evaluation}
\label{sec:evaluations-qualitative}
We compare the results of three approaches: Retrieval-Centric Generation (RCG), Retrieval-Augmented Generation (RAG), and Retrieval-OFF Generation (ROG). Note that in this work, we define RAG as allowing more permissible integration of LLM's inherent and externally retrieved knowledge, whereas RCG prioritizes clear demarcations between context interpretation and knowledge memorization. Investigating advanced methods in extracting RCG behavior is a promising research topic. In this work, we conduct simple experiments using prompt-engineering technique to reveal the potential of RCG approach. Specifically, for RCG, we employ a retrieval suffix prompt that reads ``\emph{answer the following question with the provided knowledge.}'' For RAG, we use a less constraining prompt that reads ``\emph{answer the following question. You may use the provided knowledge.}'' Finally, for ROG, which is the baseline in our evaluation, we disable the retriever altogether. By using the prompt-engineering function in \toolname{}, we can easily perform these experiments. Queries are human annotated. In Table \ref{table:prompt-engineering} and \ref{table:retrieval-centric}, blue-colored texts indicates factually correct responses while red-colored texts indicates hallucinations.

\begin{table}[hbtp]
  \small
  \centering
  Query: where is Kioxia's factory?
  \begin{tabular}{p{0.1\columnwidth}p{0.8\columnwidth}}
    \hline
    Approach & Response  \\
    \hline \hline
    RCG & \textcolor{blue}{Kioxia has factories located in both Yokkaichi city, Mie prefecture and Kitakami city, Iwate prefecture in Japan.} \\
    \hline
    RAG & \textcolor{red}{Kioxia has factories located in various regions across the globe including Japan, Thailand, Malaysia, China, and the United States. However,} \textcolor{blue}{its main production site is located in Yokkaichi City, Mie Prefecture, Japan.} \\
    \hline
    ROG & \textcolor{red}{Our main production site for HDD and SSD products is located in Yamagata, Japan. We also have manufacturing sites in Thailand and the Philippines that produce optical disc media such as Blu-ray Discs and DVDs.} \\
    \hline
  \end{tabular}
  \caption{Qualitative comparison between Retrieval-Centric, Retrieval-Augmented and Retrieval-OFF approaches using prompt-engineering in \toolname{}.}
  \label{table:prompt-engineering}
\end{table}

Based on the results in Table \ref{table:prompt-engineering}, we observed that RCG successfully answered our query accurately, whereas RAG attempted to integrate the retrieved knowledge base with LLM's own knowledge of the specific organization, resulting in partially erroneous information. Meanwhile, ROG responded to our query with LLM's preexisting mistaken knowledge, generating falsehoods. Our findings suggest that crafting clear and direct prompts, such as ``\emph{answer the given question using the provided knowledge}'', can encourage retrieval-centric behavior from the LLM, although further refined prompt-engineering may be necessary for optimal performance.

As shown in Table \ref{table:retrieval-centric}, RCG effectively addressed the query and produced accurate results. When combining RCG with a 50\% Explicit Prompt-Weighting (RCG-EPW), where only the first 50\% of retrieved knowledge base are injected into the prompt, the model generated partial responses but still maintained factual accuracy. In contrast, the pure LLM without RC enabled (ROG) provided hallucinatory responses, highlighting the importance of incorporating retrieval-centric techniques in language models for improved performance.

\begin{table}[hbtp]
  \small
  \centering
  Query: what did Kioxia do in Image Classification
  \begin{tabular}{p{0.2\columnwidth}p{0.7\columnwidth}}
    \hline
    Approach & Response  \\
    \hline \hline
    RCG & \textcolor{blue}{Kioxia developed a high capacity storage device which could store large amounts of image data required by AI algorithms. This allowed them to deploy memory centric AI systems capable of handling multiple classification tasks simultaneously without compromising performance or accuracy.} \\
    \hline
    RCG-EPW & \textcolor{blue}{Deployment of Memory-centric AI with high capacity storage} \\
    \hline
    ROG & \textcolor{blue}{The company developed a new image classification algorithm} \textcolor{red}{that achieved top-1 accuracy of 82.7\% on the ImageNet validation set, which is an improvement over previous state-of-the-art results by more than one percentage point. This was made possible through advancements in hardware and software technologies such as Tensor Core GPUs and our own deep learning framework, ONNX-MXNET.} \\
    \hline
  \end{tabular}
  \caption{Effects of Retrieval-Centric Generation in \toolname{}, based on the knowledge base about an organization.}
  \label{table:retrieval-centric}
\end{table}

\subsection{Accuracy \& Speed Evaluations}
\label{sec:evaluations-accuracy-speed}

In addition to evaluating the effectiveness of RCG using human annotations, we also conduct an internal evaluation of our tool's performance using a self-generated dataset. To create this dataset, we pass relevant passages through the language model Llama-2-13B-chat \citep{touvron2023llama} to generate 10 query and label pairs. For details on how we generated this dataset, refer to Appendix \ref{sec:evaluation-data}. We employ Rouge-L score \citep{lin-2004-rouge} as our performance metric. We perform this evaluation by using the API function of \toolname{}. Our results in Table \ref{table:response-accuracy-speed} show that RCG significantly improves the Rouge-L score compared to the baseline approach of ROG, while also slightly more competitive than RAG. Moreover, despite the fact that RCG processes longer prompts than ROG due to the addition of knowledge tokens, we observe a decrease in processing time owing to the increased precision and brevity of the generated responses. Specifically, number of response tokens generated in RCG are in average 36\% less than those generated in ROG. This efficient performance may facilitate broader adoption within the community, as users can expect quicker response generation without sacrificing accuracy.

\begin{table}[h]
  \small
  \centering
  \begin{tabular}{|c|c|c|}
    \hline
    Approach & Rouge-L Score & time/query(s) \\
    \hline \hline
    ROG & 0.186 & 17.22 \\
    \hline
    RAG & 0.359 & 18.41 \\
    \hline
    RCG & 0.413 & 11.67 \\
    \hline
  \end{tabular}
  \caption{Response accuracy \& speed evaluation of \toolname{}.}
  \label{table:response-accuracy-speed}
\end{table}

Finally, our findings suggest that even a modestly sized LLM of 13B parameters can demonstrate satisfactory performance in RCG approach towards never-seen-before factual knowledge without any model fine-tuning, potentially facilitates the deployment of Generative AI systems in real-world scenarios. See Appendix \ref{sec:applications} for further discussions and \ref{sec:ablation-study} for ablation studies.

\section{Conclusion}
\label{sec:conclusion}
We introduced \toolname{}, an open-source tool that aims to provide a localizable, lightweight, and user-friendly GUI and API platform for a Retrieval-Centric Generation approach based on LLMs. Our tool enables developers and end-users to easily interact and develop with a privacy-preserving and locally implemented LLM-based RCG system, which we believe will contribute to the democratization of these technologies within the machine learning community. Increased clarity in role-separation between context interpretation and knowledge memorization can potentially boost the performance and interpretability of generative AI systems, facilitating deployments.

\section*{Limitations}
\label{sec:limitations}
It is important to note that this tool does not provide a foolproof solution for ensuring a completely safe and responsible response from generative AI models, even within a retrieval-centric approach. The development of safer, interpretable, and responsible AI systems remains an active area of research and ongoing effort.

Generated texts from this tool may exhibit variations, even when only slightly modifying prompts or queries, due to the next token prediction behavior of current-generation LLMs. This means users may need to carefully fine-tune both the prompts and queries to obtain optimal responses.


\bibliography{custom}
\bibliographystyle{acl_natbib}

\appendix
\section{Appendix}
\label{sec:appendix}

\subsection{GUI Design of Retrieval Tuning Module}
\label{sec:gui-retrieval}

\begin{figure*}[hbtp]
	\centering
  \includegraphics[width=\textwidth]{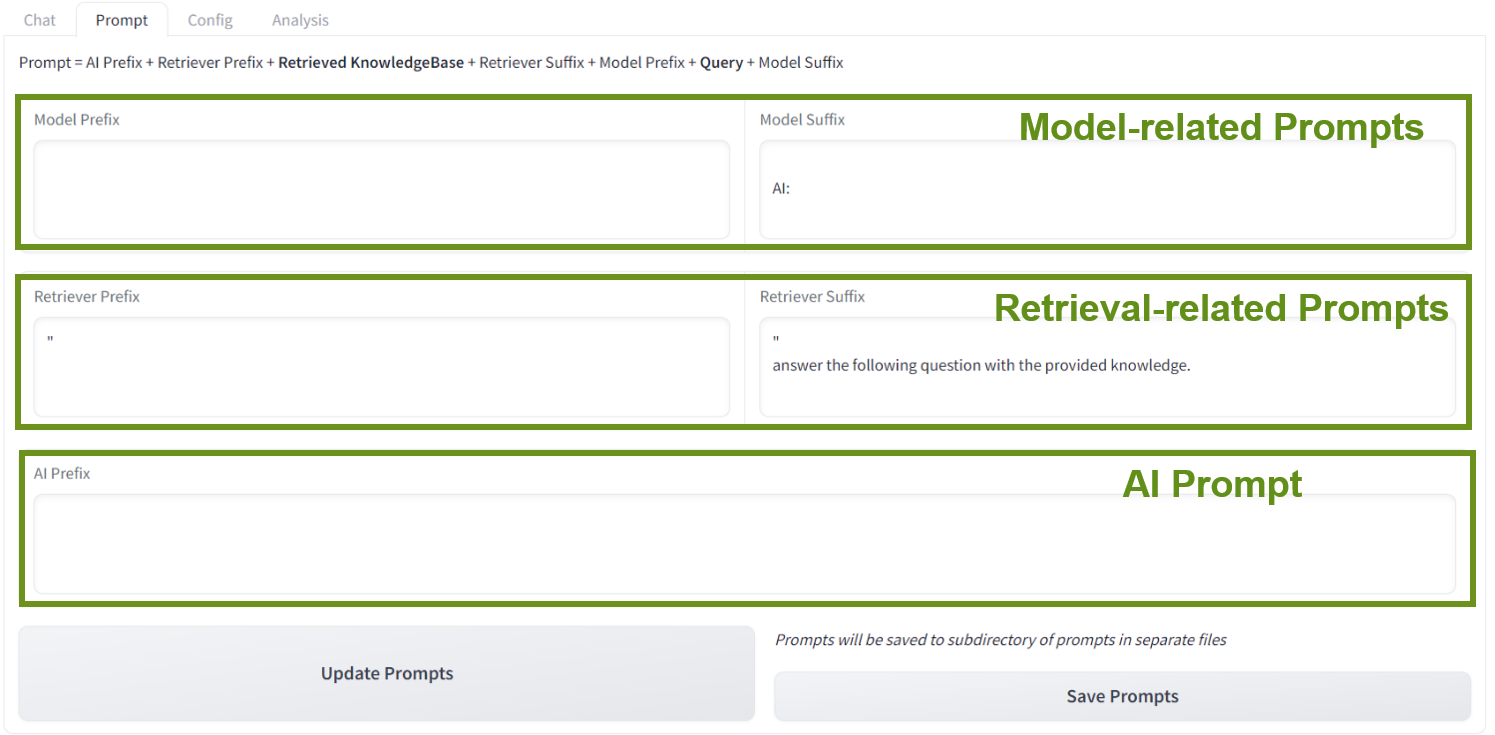}
  \caption{The Prompt-Engineering interface of \toolname{}. The Tab is for editing, updating and saving of model-related and retrieval-related prompts. Available prompts are \emph{AI Prefix, Retriever Prefix, Retriever Suffix, Model Prefix and Model Suffix}.}
	\label{Fig:guiprompt}
\end{figure*}

\begin{figure*}[hbtp]
	\centering
  \includegraphics[width=\textwidth]{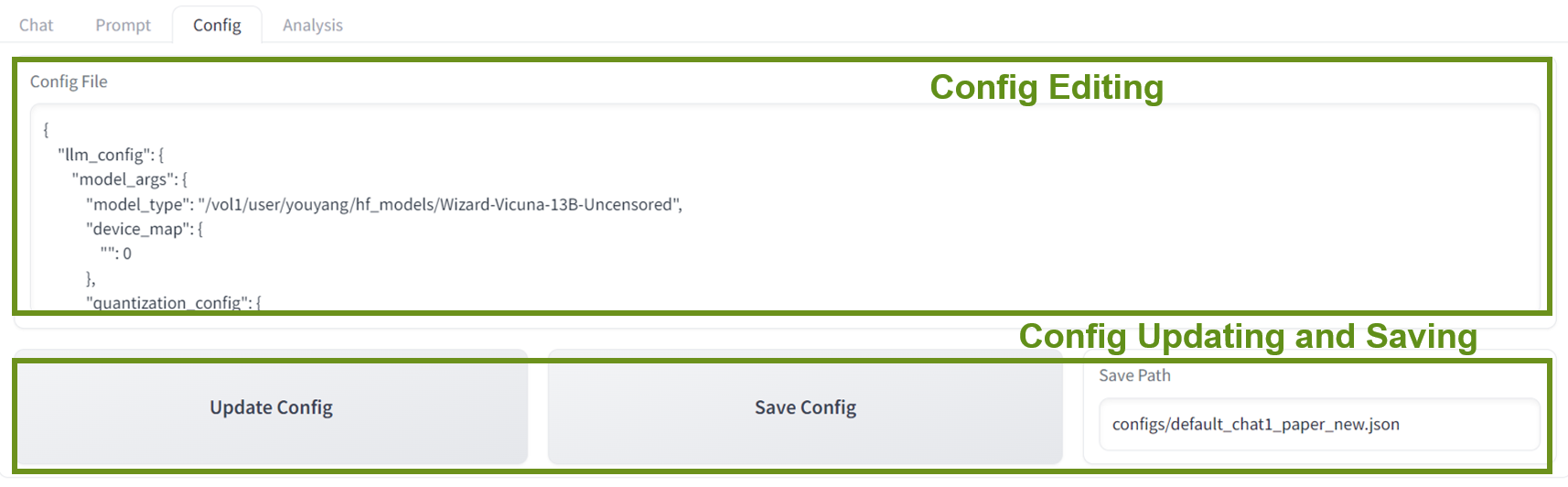}
  \caption{The Tool Configuration interface of \toolname{}. The Tab is for modifying, updating and saving all configurable settings.}
	\label{Fig:guiconfig}
\end{figure*}

\begin{figure*}[hbtp]
	\centering
  \includegraphics[width=\textwidth]{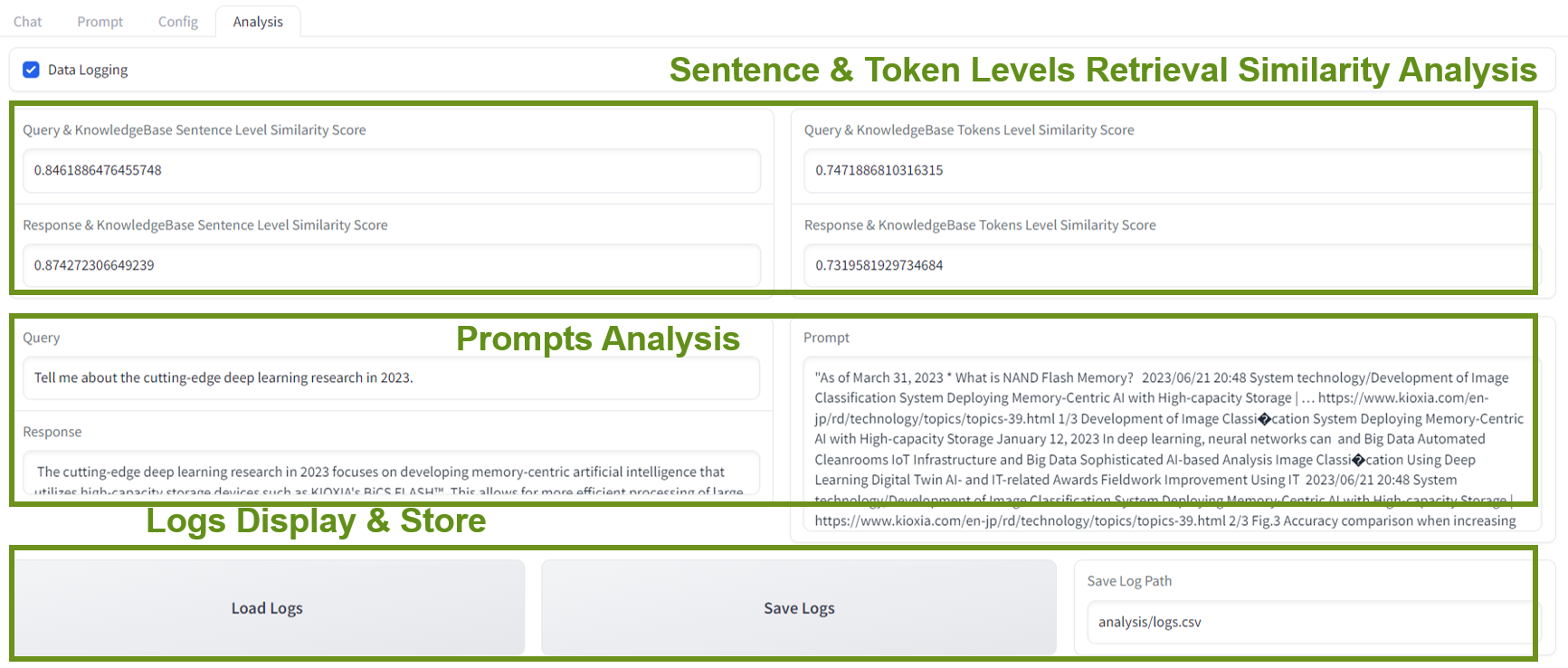}
  \caption{The Analysis and Data Logging interface of \toolname{}. The Tab is for analysis and logging of retrieved knowledge base, query, response, retrieval sentence level and tokens level similarity scores. Users can save all the logged data for further analysis.}
	\label{Fig:guianalysis}
\end{figure*}

Figure \ref{Fig:guiprompt} shows the GUI design of prompt-engineering interface.
Figure \ref{Fig:guiconfig} shows the GUI design of tool configuration interface.
Figure \ref{Fig:guianalysis} shows the GUI design of analysis and data logging interface.

\subsection{Applications}
\label{sec:applications}

\toolname{} has vast potential for various practical applications. For instance, it can serve as the foundation for building private, personalized, and lightweight generative AI systems. Sensitive and personal information can be securely stored and processed within the retrieval-centric platform. This approach enables organizations to develop interpretable and locally tailored generative AI systems for critical infrastructure. Additionally, the use of a relatively smaller language model as a contextual interpreter in this approach facilitates seamless integration into edge computing environments. The decreasing costs of data storage devices also make it feasible to establish large-scale knowledge bases. Furthermore, \toolname{} paves the way for the development of LLM-based personalized AI assistants. Lastly, an in-depth exploration of LLM-based retrieval-centric generation using \toolname{} may offer valuable insights and opportunities for future research.

\subsection{Prompt Catalogs}
\label{sec:prompt-catalogs}

Table \ref{table:prompt-evaluation} shows the prompts used in the evaluation results of Section \ref{sec:evaluations} while Table \ref{table:prompt-catalogs} shows sample prompts that may exhibit retrieval-centric behaviors. Prompts are passed to LLM in the following format: \emph{AI Prefix + Retriever Prefix + Retrieved Knowledge Base + Retriever Suffix + Model Prefix + Query + Model Suffix}.

\subsection{Evaluation Data}
\label{sec:evaluation-data}

Table \ref{table:evaluation-data} presents the data used for evaluating the performance of our proposed tool in Section \ref{sec:evaluations-accuracy-speed}. We employed the Llama-2-13B-chat model \citep{touvron2023llama} with a customized prompt (\emph{"relevant information." Please create a query and answer from the paragraph above}) to generate query and label pairs automatically from relevant information on the website of an organization.

\subsection{Ablation Study}
\label{sec:ablation-study}

As shown in Table \ref{table:ablation-study}, our ablation study reveals that adjusting Explicit Prompt-Weighting in \toolname{} leads to significant improvements in Rouge-L scores. Interestingly, increasing the weightage to 50\% yields the highest improvement, beyond which the performance remains relatively stable. This suggests that the top 50\% of retrieved knowledge bases are crucial for achieving high accuracy. However, it is important to note that these findings may not generalize to all datasets or knowledge bases, and further investigation may be necessary to determine optimal weightages for specific use cases. In comparing the response times for each query across different settings, we observe that the response times remain relatively consistent for all cases of RCG, while they increase significantly in the baseline (ROG) setting. Despite the fact that RCG processes longer prompts than the baseline, we observe a decrease in processing time owing to the increased precision and brevity of the generated responses.

\onecolumn
\begin{table*}[hbtp]
  \small
  \centering
  \begin{tabular}{|c|c|c|}
    \hline
    Approach & Rouge-L &  time/query(s) \\
    \hline \hline
    ROG & 0.186 & 17.22 \\
    \hline
    RCG-EPW-10 & 0.275 & 12.72 \\
    \hline
    RCG-EPW-20 & 0.313 & 13.00 \\
    \hline
    RCG-EPW-30 & 0.403 & 13.06 \\
    \hline
    RCG-EPW-40 & 0.354 & 11.98 \\
    \hline
    RCG-EPW-50 & 0.414 & 12.46 \\
    \hline
    RCG-EPW-60 & 0.331 & 11.36 \\
    \hline
    RCG-EPW-70 & 0.392 & 13.56 \\
    \hline
    RCG-EPW-80 & 0.306 & 16.32 \\
    \hline
    RCG-EPW-90 & 0.378 & 13.13 \\
    \hline
    RCG & 0.413 & 11.67 \\
    \hline
  \end{tabular}
  \caption{Ablation study of Explicit Prompt-Weighting in \toolname{}.}
  \label{table:ablation-study}
\end{table*}

\begin{table*}[hbtp]
  \small
  \centering
  \begin{tabularx}{\textwidth}{|l|l|l|l|l|}
    \hline
    AI Prefix & Retriever Prefix & Retriever Suffix & Model Prefix & Model Suffix \\
    \hline \hline
    & " & \makecell[Xt]{" \\ answer the following question with the provided knowledge. \\ \vphantom{ }} & & \makecell[Xt]{ \\ AI:} \\
    \hline
  \end{tabularx}
  \caption{Prompts used in the evaluation results of Section \ref{sec:evaluations}.}
  \label{table:prompt-evaluation}
\end{table*}

\begin{table*}[hbtp]
  \small
  \centering
  \begin{tabularx}{\textwidth}{|l|l|l|l|l|}
    \hline
    AI Prefix & Retriever Prefix & Retriever Suffix & Model Prefix & Model Suffix \\
    \hline \hline
    \makecell[Xt]{you are a Retrieval-Centric AI. Knowledge below are provided. \\ \vphantom{ }} & " & \makecell[Xt]{" \\ only use the provided knowledge to answer the following question. \\ \vphantom{ }} & & \makecell[Xt]{ \\ Response:} \\
    \hline
    & " & \makecell[Xt]{" \\ answer the following question with the provided knowledge. \\ \vphantom{ }} & & \makecell[Xt]{ \\ AI:} \\
    \hline
    & " & \makecell[Xt]{" \\ only use the provided knowledge to answer the following question. \\ \vphantom{ }} & & \makecell[Xt]{ \\ AI:} \\
    \hline
    \makecell[Xt]{you are a Retrieval-Centric AI. Knowledge below are provided. \\ \vphantom{ }} & " & \makecell[Xt]{" \\ only use the provided knowledge to answer the following question. \\ \vphantom{ }} & & \makecell[Xt]{ \\ AI:} \\
    \hline
  \end{tabularx}
  \caption{Sample Prompts Catalog of Retrieval-Centric Generation in \toolname{}.}
  \label{table:prompt-catalogs}
\end{table*}

\begin{table*}[hbtp]
  \small
  \centering
  \begin{tabularx}{\textwidth}{p{0.2\textwidth}p{0.7\textwidth}}
    \hline
    Query & Label \\
    \hline \hline
    What is KIOXIA known for in terms of innovation? & Based on the text, KIOXIA is known for inventing NAND flash memory in 1987 and advancing the information society on a global scale through its R\&D efforts. The company has also been recognized for presenting its research and development achievements through various channels, including major academic societies and research conferences. \\
    \hline
    What is the purpose of establishing KIOXIA Iwate Corporation? & The purpose of establishing KIOXIA Iwate Corporation is to meet the growing demand for flash memory by leveraging advanced manufacturing processes that utilize AI. \\
    \hline
    When did Kioxia start operating its new fabrication facility (Fab7)? & According to the text, Kioxia started operating its new fabrication facility (Fab7) in the fall of 2022. \\
    \hline
    What are some ways in which the Yokkaichi Plant contributes to society? & The Yokkaichi Plant contributes to society by manufacturing high-quality flash memory and SSDs that support the digitalization of various industries, as well as through its research and development (R\&D) activities. Additionally, the plant prioritizes legal compliance, environmental protection, and social initiatives, making it a responsible corporate citizen. \\
    \hline
    What is KIOXIA's approach to corporate social responsibility? & According to the passage, KIOXIA's approach to corporate social responsibility involves making their Yokkaichi Plant a factory that is admired and trusted by everyone in the community. They aim to contribute to local development and meet the expectations of local communities by being a responsible member of society. \\
    \hline
    What kind of environmental initiatives has KIOXIA taken at its Yokkaichi Plant? & At KIOXIA's Yokkaichi Plant, all employees, including resident suppliers, actively participate in various initiatives to contribute to society by improving the environment. These include measures to mitigate the environmental burden of the company's semiconductor manufacturing facilities. \\
    \hline
    What was the total revenue for the Kioxia group in fiscal year 21? & According to the passage, the total revenue for the Kioxia group in fiscal year 21 was 1,526.5 billion yen. \\
    \hline
    What initiatives has the Yokkaichi Plant taken to protect the environment? & The Yokkaichi Plant has taken various initiatives such as reducing greenhouse gas emissions, conserving water resources, and implementing recycling programs to protect the environment. These efforts aim to not only reduce the plant's ecological footprint but also raise awareness among local communities about the importance of environmental conservation. \\
    \hline
    What is KIOXIA's mission statement? & According to the passage, KIOXIA's mission statement is to uplift the world with 'memory', which means they aim to provide high-quality memory and SSD products that offer fast speeds, large capacities, and reliable performance. Additionally, the company seeks to achieve this goal through collaboration and knowledge sharing across different departments and organizations. \\
    \hline
    What is the purpose of collecting over two billion data items daily? & To maintain high-quality products by analyzing the vast amounts of data generated during the manufacturing process using advanced technologies like deep learning and AI. \\
    \hline
  \end{tabularx}
  \caption{Dataset used in the evaluation results of Section \ref{sec:evaluations-accuracy-speed}.}
  \label{table:evaluation-data}
\end{table*}
\twocolumn

\end{document}